# Let us first agree on what the term "semantics" means: An unorthodox approach to an age-old debate

Emanuel Diamant
<emanl.245@gmail.com>
http://www.vidia-mant.info

**Abstract:** Traditionally, semantics has been seen as a feature of human language. The advent of the information era has led to its widespread redefinition as an information feature. Contrary to this praxis, I define semantics as a special kind of information. Revitalizing the ideas of Bar-Hillel and Carnap I have recreated and re-established (on totally new grounds) the notion of semantics as the notion of Semantic Information. I have proposed a new definition of information (as a description, a linguistic text, a piece of a story or a tale) and a clear segregation between two different types of information – physical and semantic information. I hope, I have clearly explained the (usually obscured and mysterious) interrelations between data and physical information as well as the relation between physical information and semantic information. Consequently, usually indefinable notions of "information", "knowledge", "memory", "learning" and "semantics" have also received their suitable illumination and explanation.

## 1. Introduction

Semantics, as a facet of human language, has always attracted the attention of notable philosophers and thinkers. Wikipedia relates the first insights into semantic theory to Plato, the great Greek philosopher of the ancient times, somewhere about the end of the 4th century BC (Wikipedia, 2011). Nevertheless, despite the long history of investigations, the notion of semantics remains elusive and enigmatic. Only in the second half of the passed century (and partially on the verge of the current one) some sort of a consensual definition had emerged.

Alfred Tarski defined semantics as "a discipline which, speaking loosely, deals with certain relations between expressions of a language and the objects… 'referred to' by those expressions" (Tarski, 1944).

Jerry Fodor defines semantics as "a part of a grammar of (a) language. In particular, the part of a grammar that is concerned with the relations between symbols in the language and the things in the world that they refer to or are true of" (Fodor, 2007).

In the latest issue of "The Handbook of Computational Linguistics", David Beaver and Joey Frazee give another, slight different, definition of semantics: "Semantics is concerned with meaning: what meanings are, how meanings are assigned to words, phrases and sentences of natural and formal languages, and how meanings can be combined and used for inference and reasoning" (Beaver & Frazee, 2011).

The list of such citations can be extended endlessly. Nevertheless, an important and an interesting point must be mentioned here – the bulk of citations presuppose a tight link between semantics and the language that it is intended to work for. And that is not surprising – language was always seen as an evolutionary feature that has made us human, that is, a thing that has facilitated our ability to interact and cooperate with other conspecies. It is commonly agreed that the spoken language was the first and the ultimate tool that has endowed us (humans) with the ability to communicate, thus enormously improving our chances of survival.

But speaking about the spoken language and its role in human communication, we cannot avoid the inevitable, and somewhat provocative, question: "What actually is being communicated? The first answer which comes to mind is – language. But we have just agreed that language is only a tool that has emerged to reify our ability to communicate.

I will not bother you with rhetorical questions. My answer is simple, fair and square: Semantics – that is what we communicate to our conspecies using the language as a tool for communication.

It is perfectly right to stress that spoken language was the most ancient enabling technology evolutionary evolved for communication purposes. However, in the course of human development other means of communication have gradually emerged – cave paintings, written languages, book-printing and, in more modern times, various electrical (telegraph, telephony) and electronic (radio, television, internet) forms of communication. What were they all intended to communicate?

You will possibly reject my speculations, but I will insist – Semantics that is what we are all communicating! And it does not matter if in our modern age you prefer to call it not "Semantics" but "Information". (I hope my readers would easily agree that 13,900,000 results for a Google inquiry about "information communication" (and 2,720,000 results for "communicating information") are enough convincing to justify the claim that information is the major subject that's being communicated nowadays).

You will possibly remind me that the first attempt to integrate the terms "Semantics" and "Information" was made about 60 years ago by Yehoshua Bar-Hillel and Rudolf Carnap (Bar-Hillel & Carnap, 1952). As to my knowledge, they were the first who coined the term "Semantic Information". They have sincerely believed that such a merging can be possible: "Prevailing theory of communication (or transmission of information) deliberately neglects the semantic aspects of communication, i. e., the meaning of the messages… Instead of dealing with the information carried by letters, sound waves, and the like, we may talk about the information carried by the sentence, 'The sequence of letters (or sound waves, etc. ). .. has been transmitted' " (Bar-Hillel & Carnap, 1952).

However, they were not successful in their try to unite the mathematical theory of information and semantics. The mainstream thinking of that time was determined by the famous saying of The Mathematical Theory of Communication fathers (Claude E. Shannon and Warren Weaver): "These semantic aspects of communication are irrelevant to the engineering problem… It is important to emphasize, at the start, that we are not concerned with the meaning or the truth of messages; semantics lies outside the scope of mathematical information theory", (Shannon & Weaver, 1949).

I hope my readers are aware that denying any relations between semantics and information was not the most inspiring idea of that time. On the contrary, for many years it has hampered and derailed the process of understanding the elusive nature of them both, semantics and information alike (Two concepts that in course of human history have become the most important features of human's life).

The aim of this paper was to avoid the historical pitfalls and not to repeat the mistakes and misconceptions so proudly preached by our predecessors. I will try to prove the existence of a firm link between semantics and information and I will make my best trying to share with you my understanding of their peculiarities, which have been unveiled in course of my research into the subject of our discourse.

## 2. What is information?

The question "What is information?" is as old and controversial as the question "What is semantics?" I will not bore you with re-examining what the most prominent thinkers of our time have thought and said about the notion of "Information". In the paper's reference list I provide some examples of their viewpoints (Adams, 2003; Floridi, 2005; Sloman, 2011) with only one and a single purpose in mind – curious readers by themselves would decide how relevant and useful (for our discussion about semantics/information interrelations) these scholar opinions are.

### 2.1 Visual information, the first steps

My personal interaction with information/semantics issues has happened somewhere in the mid-1980s. At that time I was busy with home security and surveillance systems design and development. As known, such systems rely heavily on visual information acquisition and processing. However – What is visual information? – nobody knew then, nobody knows today. But, that has never restrained anybody from trying again and again to meet the challenge.

Deprived from a suitable understanding what visual information is, computer vision designers have always tried to find their inspirations in biological vision analogs, especially human vision analogs. Although underlying fundamentals and operational principles of human vision were obscure and vague, still the research in this field was always far more mature and advanced. Therefore the computer vision society has always considered human vision conjectures as the best choice to follow.

A theory of human visual information processing has been established about thirty years ago by the seminal works of David Marr (Marr, 1982), Anne Treisman (Treisman & Gelade, 1980), Irving Biederman (Biederman, 1987) and a large group of their followers. Since then it has become a classical theory, which dominates today all further developments both in human and the computer vision. The theory considers human visual information processing as an interplay of two inversely directed processing streams. One is an unsupervised, bottom-up directed process of initial image information pieces discovery and localization (The so-called low-level image processing).  The other is a supervised, top-down directed process, which conveys the rules and the knowledge that guide the linking and binding of these disjoint information pieces into perceptually meaningful image objects (The so-called high-level or cognitive image processing).

While the idea of low-level processing from the very beginning was obvious and intuitively appealing (therefore, even today the mainstream of image processing is occupied mainly with low-level pixel-oriented computations), the essence of high-level processing was always obscure, mysterious, and undefined. The classical paradigm said nothing about the roots of high-level knowledge origination or about the way it has to be incorporated into the introductory low-level processing. Until now, however, the problem was usually bypassed by capitalizing on the expert domain knowledge, adapted to each and every application case. Therefore, it is not surprising that the whole realm of image processing had been fragmented and segmented according to high-level knowledge competence of the respected domain experts.

## 2.2 Visual information and Marr's edges

Actually, the idea of initial low-level image information processing had been initially avowed by Marr (Marr, 1978). (By the way, Marr was the first who originally coined the term "visual information"). According to Marr's theory, image edges are the main bearers of visual information, and therefore, image information processing has always been occupied with edge processing duties.

Affected by the mainstream pixel-based (edge-based) bottom-up image exploration philosophy, I have at first only slightly diverged from the common practice. But slowly I have begun to pave my own way. Initially I have invented the Single Pixel Information Content measure (Diamant, 2003). Then, experimenting with it, I have discovered the Information Content Specific Density Conservation principle (Diamant, 2002). (The reverse order of publication dates does not mean nothing. Papers are published not when the relevant work is finished, but when you are lucky to meet the personal views of a tough reviewer).

The Information Content Specific Density Conservation principle says that when an image scale is successively reduced, Image Specific Information Density remains unchanged. That explains why we usually launch our observation with a general, reduced scale preview of a scene and then zoom in on the relevant scene part that we are interested in. That also indicates that we perceive our objects of interest as dimensionless items. Taking into account these observations, Information Content Specific Density Conservation phenomenon actually leads us to a conclusion that information itself is a qualitative (not a quantitative) notion with a clear smack of a narrative.

It is worth to be mentioned that similar investigations have been performed at a later time by MIT researchers (Torralba, 2009), and similar results have been attained considering the use of a reduced scale (32x32 pixels) images. However, that was done only in qualitative experiments conducted on human participants (but not as a quantitative work).

It is not surprising therefore that The Information Content Specific Density Conservation principle has inevitably led me to a conclusion that image information processing has to be done in a top-down fashion (and not bottom-up as it is usually considered). Further advances on this path supported by insights borrowed from Solomonoff's theory of Inference (Solomonoff, 1964), Kolmogorov's Complexity theory (Kolmogorov, 1965), and Chaitin's Algorithmic Information theory (Chaitin, 1966) have promptly led me to a full-blown theory of Image Information Content discovery and elucidation (Diamant, 2005).

## 2.3 Visual information, a first definition

In the mentioned above theory of Image Information Content discovery and elucidation I have proposed for the first time a preliminary definition of "What is information". In the year 2005 it had sounded as follows:

First of all, information is a description, a certain language-based description, which Kolmogorov's Complexity theory regards as a program that, being executed, trustworthy reproduces the original object. In an image, such objects are visible data structures from which an image is comprised of. So, a set of reproducible descriptions of image data structures is the information contained in an image.

The Kolmogorov's theory prescribes the way in which such descriptions must be created: At first, the most simplified and generalized structure must be described. (Recall the Occam's Razor principle: Among all hypotheses consistent with the observation choose the simplest one that is coherent with the data). Then, as the level of generalization is gradually decreased, more and more fine-grained image details (structures) become revealed and depicted. This is the second important point, which follows from the theory's pure mathematical considerations: Image information is a hierarchy of decreasing level descriptions of information details, which unfolds in a coarse-to-fine top-down manner. (Attention, please! Any bottom-up processing is not mentioned here! There is no low-level feature gathering and no feature binding!!! The only proper way for image information elicitation is a top-down coarse-to-fine way of image processing!)

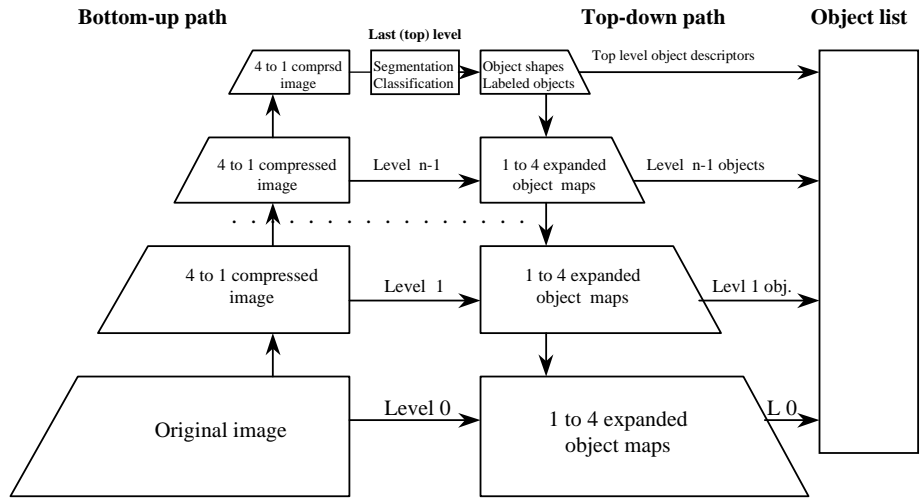

Fig. 1. The block-diagram of image contained information elucidation.

The third prominent point, which immediately pops-up from the two just mentioned above, is that the top-down manner of image information elicitation does not require incorporation of any high-level knowledge for its successful accomplishment. It is totally free from any high-level guiding rules and inspirations.

Following the given above principles, a practical algorithm for image information content discovery has been proposed and put in work. Its block-schema is provided in Fig. 1.

As it can be seen at Fig. 1, the proposed block-diagram is comprised of three main processing paths: the bottom-up processing path, the top-down processing path and a stack where the discovered information content (the generated descriptions of it) is actually accumulated.

As it follows from the schema, the input image is initially squeezed to a small size of approximately 100 pixels. The rules of this shrinking operation are very simple and fast: four non-overlapping neighbor pixels in an image at level L are averaged and the result is assigned to a pixel in a higher (L+1)-level image. Then, at the top of the shrinking pyramid, the image is segmented, and each segmented region is labeled. Since the image size at the top is significantly reduced and since in the course of the bottom-up image squeezing a severe data averaging is attained, the image segmentation/labeling procedure does not demand special computational resources. Any well-known segmentation methodology will suffice. We use our own proprietary technique that is based on a low-level (single pixel) information content evaluation (Diamant, 2003), but this is not obligatory.

From this point on, the top-down processing path is commenced. At each level, the two previously defined maps (average region intensity map and the associated label map) are expanded to the size of an image at the nearest lower level. Since the regions at different hierarchical levels do not exhibit significant changes in their characteristic intensity, the majority of newly assigned pixels are determined in a sufficiently correct manner. Only pixels at region borders and seeds of newly emerging regions may significantly deviate from the assigned values. Taking the corresponding current-level image as a reference (the left-side unsegmented image), these pixels can be easily detected and subjected to a refinement cycle. In such a manner, the process is subsequently repeated at all descending levels until the segmentation/classification of the original input image is successfully accomplished.

At every processing level, every image object-region (just recovered or an inherited one) is registered in the objects' appearance list, which is the third constituting part of the proposed scheme. The registered object parameters are the available simplified object's attributes, such as size, center-of-mass position, average object intensity and hierarchical and topological relationship within and between the objects ("sub-part of…", "at the left of…", etc.). They are sparse, general, and yet specific enough to capture the object's characteristic features in a variety of descriptive forms.

In such a way, a set of pixel clusters (segments, structures formed by nearby pixels with similar properties) is elucidated and depicted providing an explicit representation of the information contained in a given image. That means, taking the relevant segment description we can reconstruct it trustworthy and rigorously, because (by definition) every such a description contains all the information needed for the item's (or the whole set of items, that is an entire image) successful reconstruction.

## 2.4 Visual information = Physical information

One interesting thing has already been mentioned above – the top-down coarse-to-fine image information elucidation does not require any high-level knowledge incorporation for its successful accomplishment. It is totally free from any high-level guiding rules and inspirations (Which is in a striking contrast with the classic image information processing theories). It deals only with natural (physical) structures usually discernible in an image, which originate from natural aggregations of similar nearby data elements (e.g., pixels in the case of an image). That was the reason why I have decided to call it "Physical Information".

To summarize all what we have learned until now we can say:

- **Physical Information is a description of data structures** usually discernable in a data set (e.g., pixel clusters or segments in an image).

- **Physical Information is a language-based description**, according to which a reliable reconstruction of original objects (e.g., image segments) can be attained while the description is carried out (like an execution of a computer program).

- **Physical Information is a descending hierarchy** of descriptions standing for various complexity levels, a top-down coarse-to-fine evolving structure that represents different levels of information details.

- **Physical Information is** the only information that can be extracted from a raw data set (e.g., an image). Later it can be submitted to further suitable image processing, but at this stage it is the only information available.

To my own great surprise, solving the problem of physical (visual) information elucidation did not promote me even in the smallest way to my primary goal of image recognition, understanding and interpretation – they have remained elusive and unattainable as ever.

## 3. What is semantics?

### 3.1 Semantics – the physical information's twin

It is clear that physical information does not exhaust the whole visual information that we usually expect to reveal in an image. But on the other hand, it is perfectly clear that relying on our approach the only information that could be extracted from an image is the physical information, and nothing else. What immediately follows from this is that the other part of visual information, the high-level knowledge that makes grouping of disjoint image segments meaningful, is not an integral part of image information content (as it is traditionally assumed). It cannot be seen more as a natural property of an image. And it has to be seen as a property of a human observer that watches and scrutinizes an image.

This way I came to the conclusion that the notion of visual information must be disintegrated to two composite parts – physical information and semantic information. The first is contained in an image while the other is contained in the observer's head. The first can be extracted from an image while the second – and that is an eternal problem – cannot be studied by opening the human's head in order to verify its existence or to explore its peculiarities. But, if we are right in our guess that semantics is information, then we have some general principals, some insights, which can be drawn from physical information studies and applied to our semantics investigations.

In such a case, all previously defined aspects of the notion of information must also hold in the case of semantic information. That is, we can say – Semantics is a language-based description of the structures that are observable in a given image. While physical information describes structures formed by agglomeration of physical data elements (e.g., image pixels), semantic information describes structures formed by interrelations among data clusters (or segments) produced by preliminary pixel arrangements.

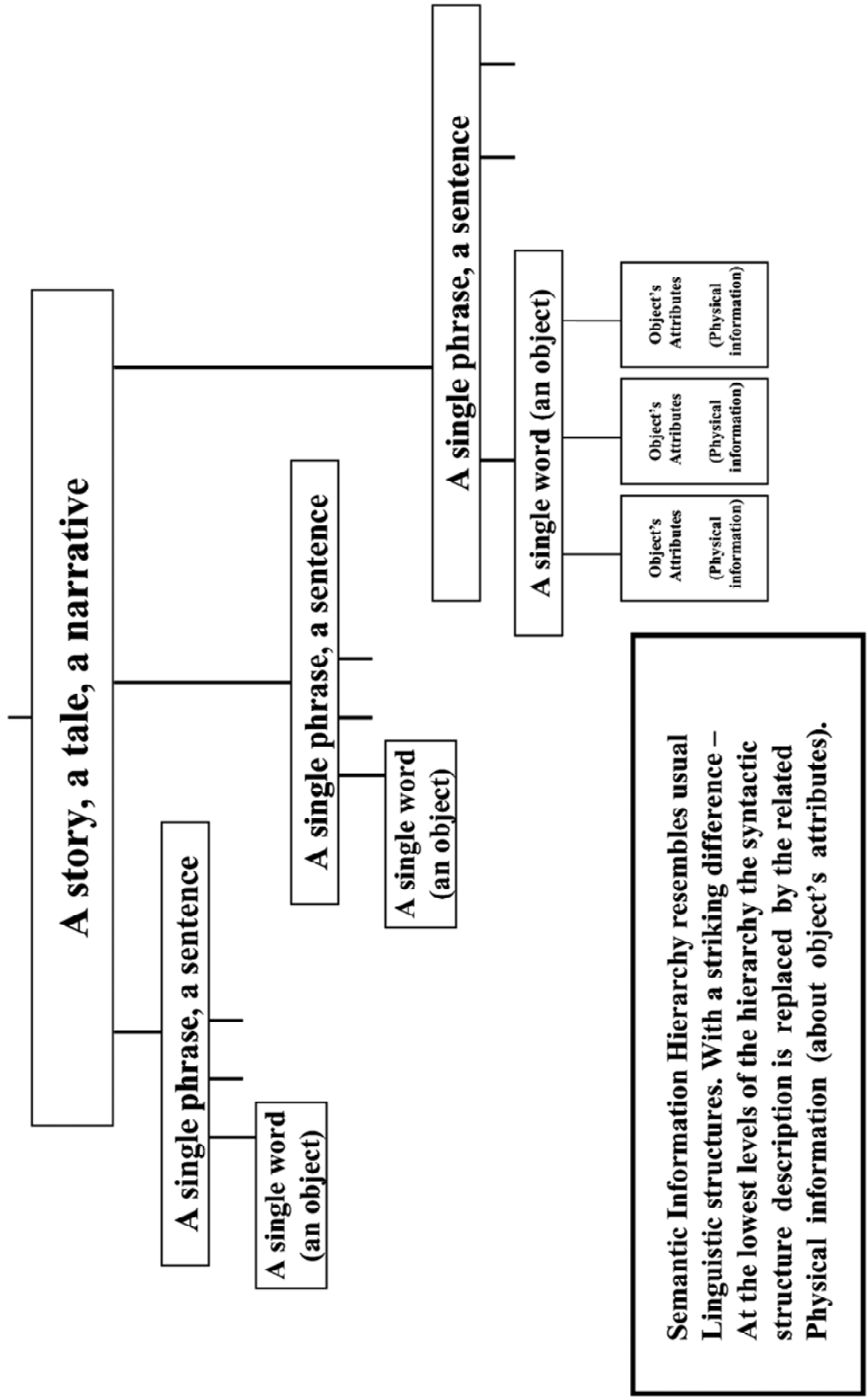

Fig. 2. Semantic Information hierarchical arrangement.

Bearing in mind this difference between semantic and physical information, we will proceed with what must be common to all information descriptions. That is, as all other information descriptions, semantics has to be a hierarchical structure which evolves in a top-down coarse-to-fine manner. Unlike physical information, which can be based on a variety of languages (be reminded – mathematics is also a sort of a language), semantic information is commonly based on the human natural language descriptions. This is the reason why historically semantics was always strongly tied with human language.

In this regard, I hope that a hypothetical semantic information hierarchical arrangement depicted in Fig. 2 can be seen as a trustworthy representation of the inner semantic information architecture. It must also be mentioned that this architecture resembles the structure of a written document, a piece of literary artwork, a tale, a paper, a narrative. As usual, it begins with a title, which is immediately followed by an abstract. From the abstract it descends to the paragraphs of the text body, and then it descends to phrases, which, in turn, are further decomposed into single separate words that build up a phrase.

## 3.2 Semantics – physical information's interpretation

At this stage, the standard linguistic semantic decomposition goes down to the syntactic components of a word. And this is not accidental – what traditional linguistics calls syntax in our case (information framework) should be called physical information (attributes), or more precisely – the underlying physical information contained in a single semantic word. That is shown in Fig. 2 as the lowest level of semantics hierarchy.

What must be specially emphasized is that these attributes (united, generalized by a higher level semantic word) could be: 1) multiple representations of different word's physical information components, which belong to the same modality; and 2) representations of different word's physical information components, which belong to different modalities. That is – the word's attributes could be represented as visual information (e.g., our case), acoustical information (as in the case of a spoken language), or any other type of physical information, including letters of a certain alphabet (as in a classical linguistic case, when a written language is used as an information bearer). It can also be a mixture of different modalities, where all physical information components are pointing (leading) to the same semantic word.

Now we can switch to the most important part of our discussion: From where semantics hierarchy does initially emerge? How it comes into existence? Somewhere in above I have mentioned that physical information is a description of structures formed by grouping of nearby data elements (I prefer to call them primary data structures) and semantic information is a description of structures formed by grouping of these nearby primary structures (I prefer to call them secondary data structures). While primary data structures are formed by grouping of nearby data elements tied by similarity in some physical property (e.g., pixel's color or brightness in an image), secondary data structures are formed without any grouping rules compliance. That means that secondary structures production (and their further naming/description) is a subjective arbitrary process, guided by mutual agreements and conventions among a specific group of observers which are involved in this (semantic) convention establishment.

This is a very important point, because what follows from it is that a new member of this specific company can not gain the established semantic conventions independently and autonomously. The conventions have to be given (transferred) to his disposal in a complete form from the outside (from the other community members) and then must be incorporated into his semantic information hierarchy. That is, fused and memorized in this hierarchy.

Publications dealing with some similar and related issues of internet documents understanding refer to this process as "a priory knowledge" acquisition and sharing. What is meant by "knowledge" is usually undefined and not considered. I think that my definition of semantic information makes it crystal clear – "Knowledge" is "Semantic information" brought from the outside and memorized into the information processing system.

And yet, the time is ripe to verify what does it means "semantic information processing"? My answer is depicted in Fig. 3 where I show how physical and semantic information are interrelated in a general information processing system.

The examination of the Fig. 3 must be prefaced with a commonplace statement that human sensory system (as well as all other so-called artificial intelligence systems) provides us only with raw sensor data and nothing else beside the data. Then, at the system's input, this sensor data is processed and physical information is being extracted from it. This physical information is fed in into the semantic information processing part, where it is matched or is being associated with the physical information stored (memorized) at the lowest level of the semantic hierarchy.

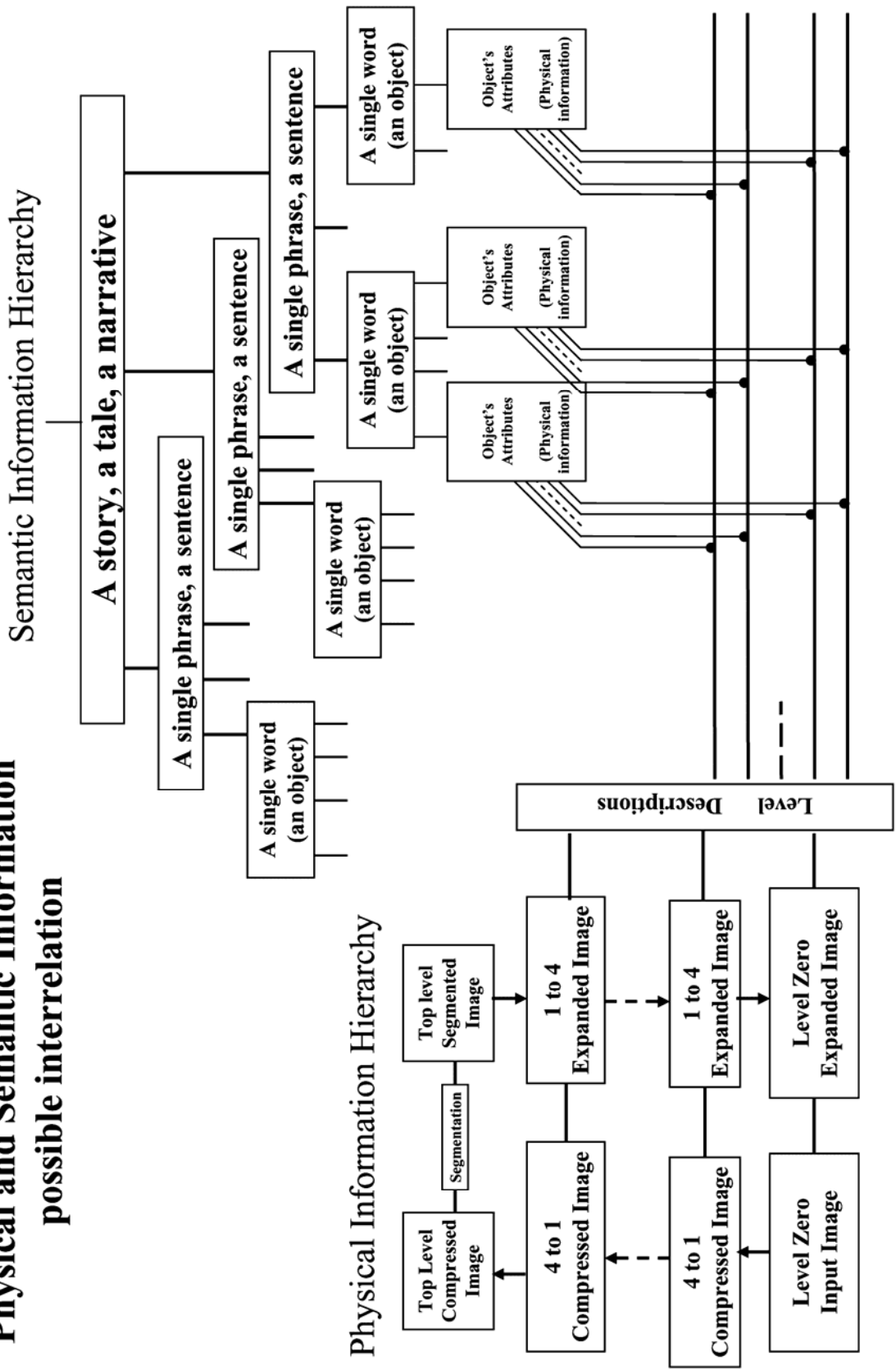

Fig. 3. Physical and Semantic Information interaction block-schema

If a match of physical information details in the input and in the stored information is attained and the details grouping conventions are satisfied, then a semantic label (an object's name) is "fired" on the first semantic level of the hierarchy. The names of adjacent objects are verified in the same manner (the so-called word's context is affirmed) and the named object finds its place in a suitable phrase or expression.

Thus the meaning of object's label (the semantics of natural language object's name) is revealed by the whole phrase in which the object (the object's name, the noun) was placed in as a suitable and a legitimate part. The semantics of a label (of a word) is defined now not only by its nearest linguistic neighbors, but by the whole phrase and the entire story text in which it is being submerged.

## 4. Generalization

Approaching the end of the paper, I would like to generalize the partial clarifications that were just given above. My research motivation was inspired by home security visual scene analysis and understanding goals. Therefore, my main concern was with visual information processing. However, I think it would be wise to broaden the scope of my findings.

I can faithfully state now that every information gathering starts with sensor data acquisition and accumulation. The body of data is not a random collection of data elements, but exhibits undeniable structures discernible in the data. These structures emerge as a result of data elements agglomeration shaped by similarity in their physical properties. Therefore, such structures could be called primary or physical data structures.

In the eyes of an external observer these primary data structures are normally grouped and tied together into more larger and complex aggregations, which could be called secondary data structures. These secondary structures reflect human observer's view on the arrangement of primary data structures, and therefore they could be called meaningful or semantic data structures. While formation of primary data structures is guided by objective (natural, physical) properties of data elements, ensuing formation of secondary structures is a subjective process guided by human habits and customs, mutual agreements and conventions.

Description of structures observable in a data set has to be called "Information". Following the given above explanation about the structures discernible in every data set, two types of information must be declared therefore – Physical Information and Semantic Information. They are both language-based descriptions; however, physical information can be described with a variety of languages, while semantic information can be described only with natural human language used.

Every information description is a top-down evolving coarse-to-fine hierarchy of descriptions representing various levels of description complexity (various levels of description details). Physical information hierarchy is located at the lowest level of the semantic hierarchy. The process of data interpretation is reified as a process of physical information extraction from the input data, followed by a process of input physical information association with physical information stored at the lowest level of a semantic hierarchy. In this way, input physical information becomes named with an appropriate linguistic label and framed into a suitable linguistic phrase (and further – in a story, a tale, a narrative), which provides the desired meaning for the input physical information.

## 5. Conclusions

In this paper I have proposed a new definition of information (as a description, a linguistic text, a piece of a story or a tale) and a clear segregation between two different types of information – physical and semantic information. I hope, I have clearly explained the (usually obscured and mysterious) interrelations between data and physical information as well as the relations between physical information and semantic information. Consequently, usually indefinable notions of "knowledge", "memory" and "learning" have also received their suitable illumination and explanation.

Traditionally, semantics is seen as a feature of human language communication praxis. However, the explosive growth of communication technologies (different from the original language-based communication) has led to an enormous diversification of matters which are being communicated today – audio and visual content, scientific and commercial, military and medical health care information. All of them certainly bear their own non-linguistic semantics (Pratikakis et al, 2011). Therefore, attempts to explain and to clarify these new forms of semantics are permanently undertaken, aimed to develop tools and services which would enable to handle this communication traffic in a reasonable and meaningful manner. In the reference list I provide some examples of such undertakings: "The Semantics of Semantics" (Petrie, 2009), "Semantics of the Semantic Web" (Sheth et al, 2005), "Geospatial Semantics" (Di Donato, 2010), "Semantics in the Semantic Web" (Almeida et al, 2011).

What is common to all those attempts is that notions of data, information, knowledge and semantics are interchanged and swapped generously, without any second thought about what implications might follow from that. In this regard, even a special notion of Data Semantics was introduced (Sheth, 1995) and European Commission and DARPA are pushing research programs aimed on extracting meaning and purpose from bursts of sensor data (Examples of such research roadmaps could be found in my last presentation at The 3-rd Israeli Conference on Robotics, November 2010, available at my website http://www.vidia-mant.info). However, as my present definition claims – data and information are not interchangeable, physical information is not a substitute for semantic information, and data is semantics devoid (semantics is a property of a human observer, not a property of the data).

Contrary to the widespread praxis (Zins, 2007), I have defined semantics as a special kind of information. Revitalizing the ideas of Bar-Hillel and Carnap (Bar-Hillel & Carnap, 1952) I have recreated and re-established (on totally new grounds) the notion of semantics as the notion of Semantic Information.

Considering the crucial role that information usage, search for, exchange and exploration have gained in our society, I dare to think that clarifying the notion of semantic information will illuminate many shady paths which Semantic Web designers and promoters are forced to take today, deprived from a proper understanding of semantic information peculiarities. As a result, information processing principles are substituted by data processing tenets (Chignell & Kealey, 2010); the objective nature of physical information is confused with the subjective nature of semantic information. For that reason, Semantic Web search engines are continue to be built relying on statistics of linguistic features. I hope my paper will let to avoid such lapses.